\documentclass{article}

\usepackage{arxiv}

\usepackage[utf8]{inputenc} 
\usepackage[T1]{fontenc}    
\usepackage{hyperref}       
\usepackage{url}            
\usepackage{booktabs}       
\usepackage{amsfonts}       
\usepackage{nicefrac}       
\usepackage{microtype}      
\usepackage{lipsum}
\usepackage{graphicx}
\usepackage{natbib}
\setcitestyle{authoryear,open={(},close={)}}

\title{Doc2Im: document to image conversion through self-attentive embedding}

\author{
  Mithun Das Gupta  \\
  Microsoft AI \& R\\
  Hyderabad, India\\
  \texttt{migupta@microsoft.com} \\
}

\begin{document}
\maketitle

\begin{abstract}
Text classification is a fundamental task in NLP applications. Latest research in this field has largely been divided into two major sub-fields. Learning representations is one sub-field and learning deeper models, both sequential and convolutional, which again connects back to the representation is the other side. We posit the idea that the stronger the representation is, the simpler classifier models are needed to achieve higher performance. In this paper we propose a completely novel direction to text classification research, wherein we convert text to a representation very similar to images, such that any deep network able to handle images is equally able to handle text. We take a deeper look at the representation of documents as an image and subsequently utilize very simple convolution based models taken as is from computer vision domain. This image can be cropped, re-scaled, re-sampled and augmented just like any other image to work with most of the state-of-the-art large convolution based models which have been designed to handle large image datasets. We show impressive results with some of the latest benchmarks in the related fields. We perform transfer learning experiments, both from text to text domain and also from image to text domain. We believe this is a paradigm shift from the way document understanding and text classification has been traditionally done, and will drive numerous novel research ideas in the community.
\end{abstract}


\section{Introduction}
Written text in form of documents, reports, emails, articles etc. forms an integral part of our everyday life. Classification, clustering, analysis and other related machine learning (ML) tasks are important components to make the huge amounts of text useful for higher level semantic tasks such a web search, sentiment analysis, open domain question answering etc. Natural language processing (NLP) and the entailing body of research also forms a complementary space which can gain by all the development in the text processing field, and vice-versa. Text and NLP based ML research was largely focused on intelligent feature design till the advent of the the deluge of deep learning. Computer vision (CV) took the pole position in this revolution, but text and NLP are very strongly catching upto these fields.
Present day text understanding and classification techniques either rely on human designed features~\cite{WangNManning2012} 
or use deep neural networks on distributed representation of texts~\cite{Conneau2016}. 
Another complementary field which is developing recently is the handling of short text snippets (\cite{Severyn:2015:}), similar to those issued as web search queries or those appearing as titles of documents and web pages. Techniques which have worked for larger text understanding seem to not work well for short text scenarios. Unlike paragraphs or documents, short texts do not always observe the syntax of natural language. Short texts usually lack any context. Short texts can also be fairly ambiguous, even to human observers, because they contain polysemes and typos. 

Sentence pair processing (\cite{ABCNN2016}) is the third related field which is largely based on classifying not one but multiple textual units at the same time. In this paper we study these varied research sub fields under one domain, with the assumption that a strong representation can lead to the adoption of standardized convolution neural network based models from CV in the text and NLP domain.


Representation models can further be divided into two categories: explicit representation and implicit representation~\cite{WangNWang2016}. 
For explicit approaches, a given text is modeled following traditional NLP steps, including chunking, labeling, and syntactic parsing. 
Although explicit models are easily understandable by human beings, it is highly prone to getting entangled into ambiguities inherent to human language. 
Additionally, they also suffers from the data sparsity problem. For example, when an entity is missing in a knowledge base, one cannot obtain any representative feature for it and forceful smoothing techniques need to be employed. 

For implicit representation, the text is represented using Neural Language Model (NLM)~\cite{Bengio2003}. 
An NLM maps texts to an implicit semantic space and parameterizes them as a vector. An implicit representation model can capture richer information from context and facilitate text understanding with the help of deep neural networks. However, apart from being non human interpretable, they also suffer from data distribution problems such as handling rare words and phrases. Representing the words in a text snippet by their word embeddings and then treating the system as a matrix on which to run CNNs was proposed by numerous authors~(\cite{Collobert:2011,kim2014}).  

In this work we propose a novel representation for text blocks, namely sentences, paragraphs and documents which can be obtained by encoding the interactions of each word with its neighbors in the text block. More specifically,
\begin{itemize}
\item We propose a very simple yet elegant representation for a text block and show that this representation is better suited for additive convolution kernels.
\item Coupled with this elegant representation, we use very simple convolution networks, lifted as is from the computer vision domain to obtain state of the art results on multiple text classification benchmark datasets.
\item Propose simple transfer learning experiments both from text to text and image to text domain. To the best of our knowledge these experiments have never been reported, at least in their most advanced sense.
\end{itemize}


\section{Related Work} 
The idea that at least some aspects of word meaning can be induced from
patterns of word co-occurrence is becoming increasingly popular. The success of word embeddings in numerous text and NLP tasks is largely due to this intuition itself. A systematic
exploration of the principal computational possibilities for formulating and validating
representations of word meanings from word co-occurrence statistics was presented by~\cite{Bullinaria2007}. 
Pairwise word interaction as a possible representation unit was proposed by~\cite{HeL16}. They proposed a pairwise word interaction model and further tied it to a similarity focus
mechanism to identify important correspondences for better similarity measurement. Although the starting point is very similar to our work, but the focus mechanism is largely hand crafted leaving the method extremely fine tuned to one particular task. 

Another work which closely follows the ideas we discuss in this work is the neural attention based sentence summarization work by~\cite{Rush2015}. 
They propose an attention framework to identify the abstractive summary for a given text block. The cost function is an NLM model with the input pair encoding being done by a simple attention mechanism. Again the simple attention mechanism comes close to the our work, but the NLM model makes the representation highly task specific.  

We propose a simple model which is both simple to generate and train. The trained model can further be used on a completely different dataset. These two facts lead to a generic concept of handling text blocks in a manner almost similar to the transfer learning scenarios which have become popular in the CV domain. One of the earliest attempts at transfer learning for text classification was proposed by~\cite{Do2005}. They proposed to use hand crafted features to represent the documents and then learned a softmax classifier. Even though this satisfies the explicit requirements for transfer learning, but the features are again tuned to a specific category of problems. Specific reference to TFIDF type of features lead to failures in generalization capabilities when we switch to modern day datasets arising out of web queries and tweets. Another similar work was proposed by ~\cite{Raina2006}, where a smoothed word correlation matrix was learned for transfer learning.

\section{Model}
We propose a document to matrix conversion by utilizing the inherent redundancies present in text. 
Word embedding methods such as GloVe proposed by~\cite{pennington2014}, use a small window around each word to define a context for the center word. This process is then iterated, whereby the inner product of the center word with the context words are maximized to generate stable representations for the words in vector spaces. This method of finding vector representations has been shown to have multiple desirable properties.

Looking at the trained embeddings, one might then assume that the embedding for a particular word is highly conditioned on the immediate neighborhood that it appears in. This essentially means a word and its immediate context neighbors span a low dimensional manifold. Locally linear embedding (LLE)~\cite{Roweis2000}, starts by finding an estimate for the low dimensional manifold and then follows it up with an eigen-decomposition step which finds the lower dimensional representation for the data. Incidentally, for very high dimensional data finding the lower intrinsic dimension to project the data itself is a challenge as shown by~\cite{Gupta2010}. But the first step of representing the center point as a linear combination of the neighboring points does indeed create a low dimensional representation for the center point to lie in. 

Another tangential way to motivate our model is self attention. Let us assume we want to find the self attention of each word in a document with its neighbors and repeat it for the entire document. 
This amounts to identifying the strong affinities present in the words constituting the document. Note that most question answering (QA) techniques such as BiDaf~(\cite{Seo2016BiDaf}) and r-net~(\cite{Wang2017RNet}), proposed to solve SQuAD dataset challenge proposed by~\cite{RajpurkarJL18}, use this attention matrix between the question and answer as the starting point for identifying the answer span. \\

\subsection{Convergence Analysis}
Let a piece of text be defined as an ordered collection of words $\{t_1, t_2, \ldots, t_N\}$. Defining the embedding for each word as $embed(t_i) = x_i$ and taking a dense inner product of all the words in the document with all other words in the same document we can write the dense self-attention matrix as
\begin{equation}
\mathbf{S_D} = [x_{ij}] ~~~ \forall ~i, ~j ~\in ~N
\end{equation}
where $x_{ij}=x_i^Tx_j$ and N is the total number of words in the document. The embedding for each word $x_i$ can be obtained from any of the state-of-the-art embedding techniques. For our comparative experiments we use the 300 dimensional GloVe embedding by~\cite{pennington2014}. 
Setting the neighborhood of every word as the entire document is highly redundant, and hence we set the neighborhood as a small interval around each word, denoted by $2k$ which is a hyper-parameter in our work. Introducing this hyper-parameter the representation for a document can now be written as
\begin{equation}
\mathbf{S_D} = [x_{ij}] ~~~ \forall ~i \in N, ~j ~\in [i-k, i+k], ~i\neq j
\end{equation}
Note that for the ends of the document, where the one sided neighborhood is smaller than $k$ we can use zero padding or we can also embed with respect to circular neighbors formed by joining the beginning to the end of the document. We call this representation as the document to image (D2I) representation. 
Starting from the first reconstruction equation of LLE by~\cite{Roweis2000}, we can write 
\begin{equation}
\epsilon(W,X) = \sum_i\|x_i - \sum_j W_{ij} x_j\|^2 
\end{equation}
where $\epsilon(W,X)$ denotes the cost of reconstructing each $x_i$ from its neighbors $x_j$'s with $W$ as the matrix of reconstruction weights.
This equation is solved individually for each $x_i$ and hence we can look into each component of the sum separately, dropping the dependence on $W$ and $X$ for notational simplicity, as:
\begin{eqnarray}
\epsilon_i &=& \epsilon(W,x_i) = \|x_i - \sum_j W_{ij} x_j\|^2 \\
x_i^T \epsilon_i x_i &=& \|x_i^T x_i - \sum_j W_{ij} x_i^T x_j\|^2
\end{eqnarray}
One of the properties of the right hand side of the reconstruction equation is the fact that it is scaling and rotation independent (\cite{Roweis2000}). Donating $x_i^T x_j$ as $x_{ij}$, we can scale the entire equation by $x_{ii}$ and write
\begin{equation}\label{Eq:1VarFormulation}
\hat{\epsilon}_i = \frac{\epsilon_i}{x_{ii}} = \|1 - \sum_j \hat{W}_{ij} x_{ij}\|^2 
\end{equation}
where we have absorbed $W_{ij}/x_{ii}$ into the new unknown weights $\hat{W}_{ij}$. One interesting way to look at Eq.~\ref{Eq:1VarFormulation} is that $\hat{W}_{ij}$ and $x_{ij}$ are entirely interchangeable in it.

Now let us look at the cost function used to learn the GloVe embedding vectors. GloVe optimization function is written as
\begin{equation}
J = \sum_{i,j=1}^V f(X_{ij})(x_{ij} + b_{ij} - \log(X_{ij}))^2
\end{equation}
where we club the two bias terms $b_i$ and $b_j$ into one joint term $b_{ij}$, $x_{ij} = x_i^T x_j$ similar to as defined earlier and $X_{ij}$ is the count of the two words $t_i$ and $t_j$ appearing in the neighborhood of each other in the entire training corpus.
Assuming the vectors have been trained, we can again separate the equation out into separate optimizations functions for each center word leading to
\begin{equation}
J_i^o = \sum_{j=1}^V f(X_{ij})(x_{ij} + b_{ij} - \log(X_{ij}))^2
\end{equation}
where $J_i^o$ denotes the optimized cost function for the center word $t_i$ at index $i$. The local optima at this point corresponds to the derivative with respect to the weight vectors being close to zero. Also note that the function $f(X_{ij})$ scales low frequency pairs, but seals the high frequency pairs to 1. Using this sealing property, at least for the high frequency pairs, we can ignore the scaling and write
\begin{equation}
J_i^o = \sum_{j=1}^V (1 - \frac{1}{\log(X_{ij})-b_{ij}}x_{ij})^2  (\log(X_{ij}) - b_{ij})^2
\end{equation}
Noting that at the local optima, the cost function changes much more slowly with respect to the bias terms $b_{ij}$ than $x_{ij}$, we can approximate the term outside the first bracket as a constant for the center word and write the final equation as
\begin{equation} \label{Eq:GloVeCost}
J_i^o/C_i =  \sum_{j=1}^V\|1 - W_{ij}^Gx_i^T x_j\|^2  
\end{equation}
where $C_i \approx \sum_j (\log(X_{ij} - b_{ij})^2 $.
This equation is a a scaled version of Eq.~\ref{Eq:1VarFormulation}, with $W_{ij}^G = \frac{1}{\log(X_{ij})-b_{ij}}$ and the superscript $G$ denotes that it is the weight equivalent for GloVe formulation. Finally by applying the linearity of expectation\footnote{http://www.math.mcgill.ca/dstephens/556/Handouts/Math556-05-Inequalities.pdf} and the fact that $E[P^2] \geq \{E[P]^2\}$, 
combining Eq.~\ref{Eq:1VarFormulation} and Eq.~\ref{Eq:GloVeCost}, we get  
\begin{equation}
J_i^o/C_i \geq \hat{\epsilon}_i
\end{equation}
and hence the GloVe cost function is a variational upper bound on the LLE reconstruction cost for high frequency neighborhoods and minimizing it leads to a neighborhood which respects the LLE constraint.\\

\subsection{Implementation Details}
\begin{figure}
\centering
\includegraphics[width=2cm]{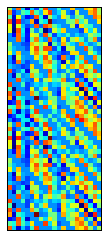}
\includegraphics[width=2cm]{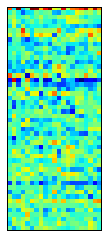}
\includegraphics[width=2cm]{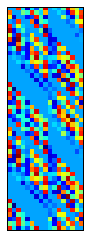}
\includegraphics[width=2cm]{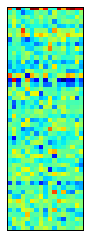}
\caption{Images generated from training examples in TwitterPPDB data. The bluish images, first and third from left are the D2I representation  with $k=25$ proposed in this work. The greenish images, second and fourth are obtained by stacking 50 dimensional GloVe vectors for all the words.}
\label{Fig:DocIm}
\end{figure}

\begin{figure}
\centering
\includegraphics[width=8cm]{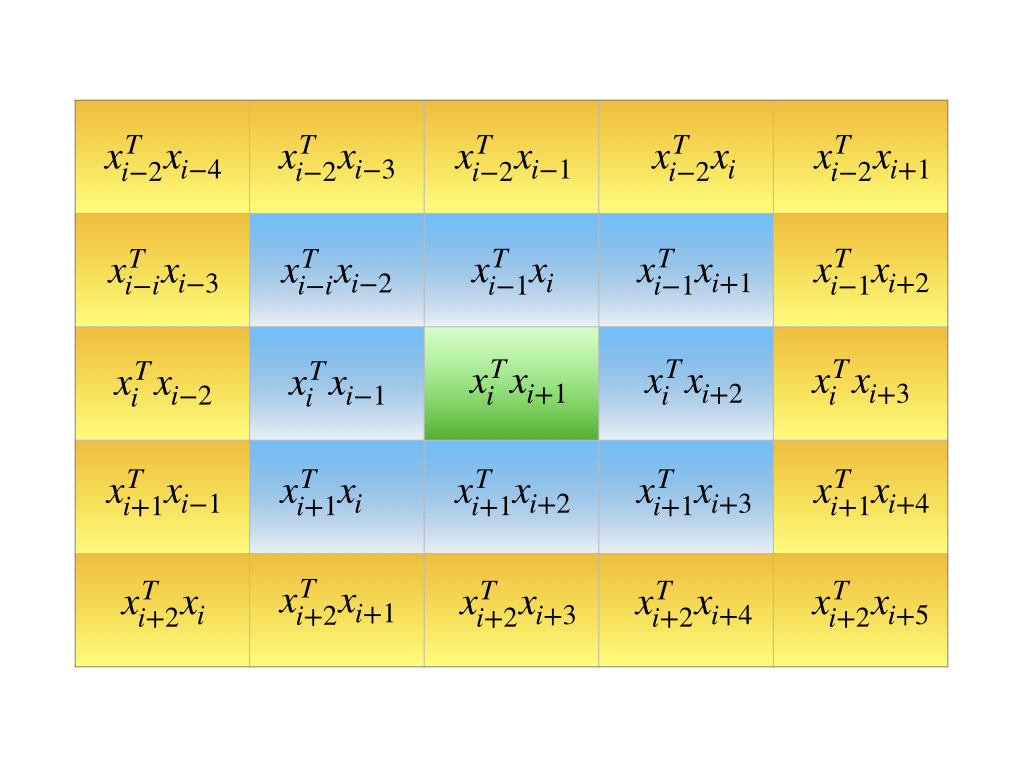}
\caption{Neighborhood structure for the proposed document to image transformation. $x_i$ is the center word, and the other words are its neighbors.}
\label{Fig:patch}
\end{figure}

Once the document has been transformed to this matrix of size $[2k, N]$ this can now be reshaped, resized and put through all the transformations which are done over matrices, more specifically images. Any unknown word, which appears as $<$UNK$>$ in many of these word embedding techniques, will appear as a row of all zeros in the document image. In our method we remove all such rows from the image. Two examples from our dataset are shown in Fig.~\ref{Fig:DocIm} (first and third panels). The diagonal patterns of solid blues refer to the unknown words which lead to zeros in the image. These images encode both the short as well as long distance correlations amongst the words in a document. For every word we explicitly encode its interactions with its $2k$ neighbors. The nearby rows encode the nearby words. If this image were to be convolved by a kernel of size $q$, then the encoding distance increases to $2k + q$. This renders a simplified way to encode long distance relationships amongst the words of a document.
Looking closely at the neighbor hood of one pixel in the new representation image, as shown in Fig.~\ref{Fig:patch}, we can now define interesting interpretations for the edges in the representation image. For the center pixel, $x_i^T x_{i+1}$, the vertical edge towards right is $x_i^T (x_{i+2} - x_{i+1})$. Similarly, the horizontal edge coming down is denoted by $x_{i+1}^T (x_{i+2} - x_{i})$, which is essentially the convolution at $x_{i+1}$ with the edge operator $[-1 ~0 ~1]^T$ multiplied by the embedding at $x_{i+1}$ to project the embeddings to the inner product space. Similarly, all the edges which can be obtained by subtracting the center pixel in green with its immediate neighbors in blue can all be represented by a convolution with a simple edge filter of type $[-1 ~0 ~1]$ followed by a projection to inner product space.

This brings us to the comparison with the preferred way of using word embeddings, which is just representing each word with its embedding and stacking them next to each other, and then applying convolution filters as proposed by numerous authors (\cite{Qiu2015,Severyn:2015:}). The two example documents from Fig.~\ref{Fig:DocIm} first and third panels from left, when represented as a matrix of their stacked word embeddings appear as shown in Fig.~\ref{Fig:DocIm} (second and fourth panels). Note that these images still show a lot more cross pollination amongst dimensions, because we train smaller 50 dimensional embeddings for representational purposes. The actual 300 dimension GloVe embedding shows extremely little cross pollination along the horizontal axis.

This representation has been used by numerous authors to create intermediate representation for sentences, documents etc. or as input to more complex models such as those proposed by~\cite{Das2016,kim2014,Collobert:2011}. 
These operations do encode correlations between the different dimensions of the representation of the same word (1D interactions) well, but do not exploit the redundancies present within the document itself. The relations between the entries of different word vectors have been shown to be fairly non-correlated\footnote{https://nlp.stanford.edu/projects/glove/}, and hence the information encoded by convolution kernels is not very well understood in this case. The banded structure of the word embeddings result from the fact that the multiplicative interactions in the GloVe model occur component-wise. While there are additive interactions resulting from the dot product in the cost function, in general there is little room for the individual dimensions to cross-pollinate. Hence any additive kernel being convolved with such a representation similar to the work by~\cite{Qiu2015} does not encode the proper information content of the document and tries to force fit an additive response from non-correlated entities. 

\begin{figure}
\centering
\includegraphics[width=8cm]{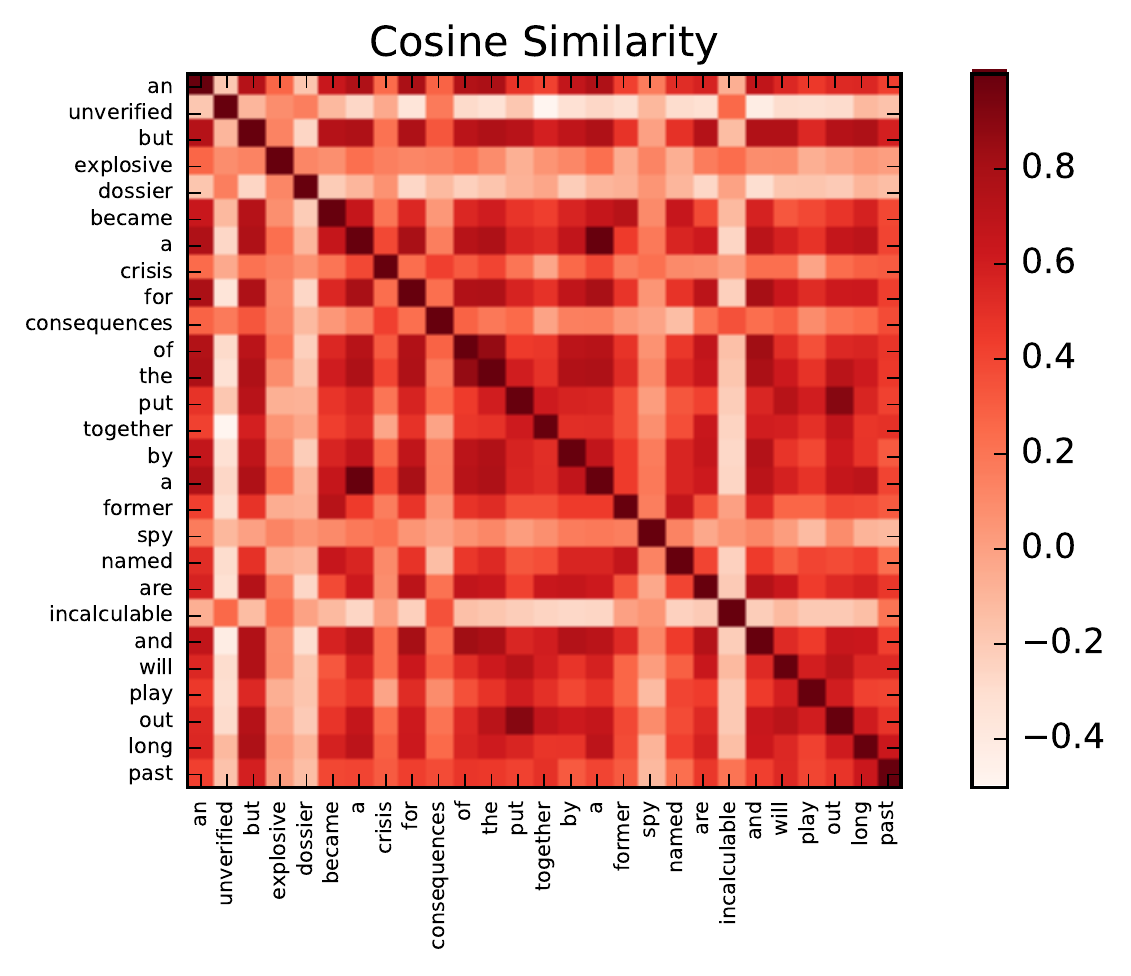}
\caption{Self attention for one of the documents as in Fig.~\ref{Fig:DocIm}.}
\label{Fig:selfAtt}
\end{figure}

Once we move to the inner product based self-attention space, these limitations are removed. This is mostly evident from the fact that in all our experiments we use the simplest convolutional neural network for MNIST as provided in the tensorflow tutorial\footnote{https://www.tensorflow.org/tutorials/estimators/cnn}. Visually it appears that the self-attention based representation encodes more information into a compact space. 
The self attention matrix for one of the example documents in Fig.~\ref{Fig:DocIm} is shown in Fig.~\ref{Fig:selfAtt}. The deep red color at the principal diagonal refers to the inner product of the word with itself. This value is removed from the representation. 
Also note that the regions near the principal diagonal have on average higher energy (darker color), with some strong low energy (lighter color) words. These regions are created by non-connected, rare words such as \textit{dossier, incalculable}. 
This also points to a lookup based refining of the self-attention image, where a rare word can be replaced by its more frequently used synonym by using a resource like WordNet (\cite{WrdNet1995}), but this has been left as a future research direction in this work.

\section{Experiments and Results}
\begin{figure*}
\centering
\includegraphics[width=16cm,height=6cm]{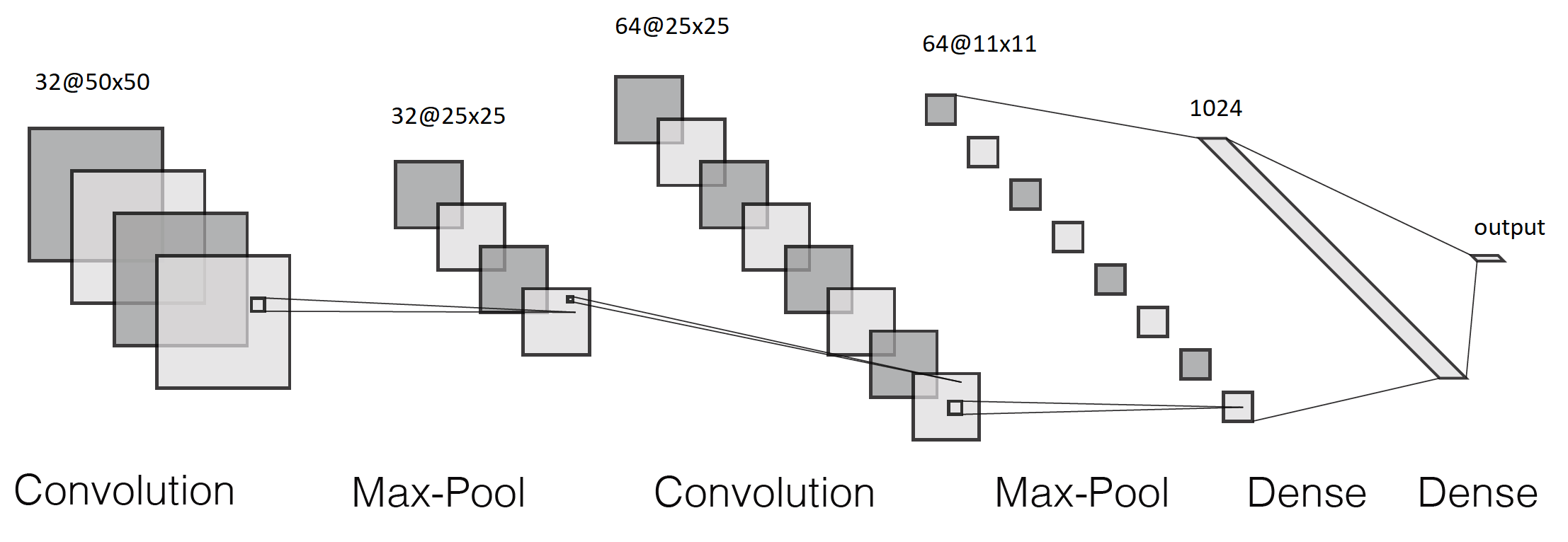}
\caption{Convolution Neural Net used for the experiments reported.}
\label{Fig:CNN}
\end{figure*}

For the experiments reported in this section we adopt a simple five layer CNN with an additional softmax layer as shown in Fig.~\ref{Fig:CNN}.
Note that the document to image (D2I) transformation remains same for all the datasets. We fix the hyper-parameter $k=[25]$. 

\subsection{Datasets}
\subsubsection{TwitterPPDB}
The TwitterPPDB\footnote{https://github.com/castorini/data/tree/master/twitterPPDB} dataset is a new dataset, which is also the largest human-labeled paraphrase corpus to date. It consists of 51,524 sentence pairs and the first cross-domain benchmarking for automatic paraphrase identification, textual similarity, and also question answering tasks. The comparative results for this dataset are shown in Table.~\ref{tab:TPPDBresults}.
\begin{table}
	\centering
\begin{tabular}{l*{2}{c}}
Method &  & Metric=F1   \\
\hline \\
Random &  & 0.3270 \\
Edit Distance &  & 0.5260 \\
DeepPairwiseWord & \cite{HeL16} & 0.7490 \\
D2I & & \textbf{0.8110} \\
\end{tabular}
	\caption{Comparative results For TwitterPPDB dataset.}
	\label{tab:TPPDBresults}
\end{table}



\subsubsection{WikiQA}
We present comparative evaluation on WikiQA~(\cite{wikiqa2015}), an open domain question-answer dataset. We use the subtask that assumes that there exists at least one correct answer for each question. The WikiQA dataset consists of 20,360 question candidate pairs for training, 1,130 pairs in validation set and 2,352 pairs in the test set, where we adopt the standard setup of only considering questions  with correct answers in test. For binary classification mean average precision is not a good metric and mean reciprocal rank for non-multilabel problems can be obtained by using label ranking average precision\footnote{http://scikitlearn.org/stable/modules/model\_evaluation.html\#label-ranking-average-precision}. The comparative results for the WikiQA dataset are shown in Table.~\ref{tab:WikiQA}. The ABCNN model proposed by~\cite{ABCNN2016} proposed a dense attention matrix generation between the question and answer representation and showed impressive gains for the WikiQA dataset. The self embedding matrix shown in Fig.~\ref{Fig:selfAtt} is quite similar to the inner product based attention proposed by~\cite{ABCNN2016}. The major difference is that we utilize the word embeddings directly to arrive at the self-attention, whereas~\cite{ABCNN2016} use convolutions over the stacked embedding representation and then generate the intermediate representations which are used to get the attention matrix. 
\begin{table}
	\centering
\begin{tabular}{l*{2}{c}}
Method &  & Metric=MRR   \\
\hline \\
Cnn-Cnt &  \cite{YuHBP14}& 0.6652 \\
ABCNN & \cite{ABCNN2016} & 0.7127 \\
D2I & & \textbf{0.9762} \\
\end{tabular}
	\caption{Comparative results For WikiQA dataset.}
	\label{tab:WikiQA}
\end{table}

\subsubsection{TrecQA}
The TrecQA dataset~(\cite{Voorhees2001}) 
from the Text Retrieval Conferences has been widely
used for the answer selection task during the past
decade. To enable direct comparison with previous
work, we used the same training, development,
and test sets as released by~\cite{Yao13}. 
The TrecQA data consists of 1,229 questions with
53,417 question-answer pairs in the TRAIN-ALL
training set, 82 questions with 1,148 pairs in the
development set, and 100 questions with 1,517
pairs in the test set. The comparative results for the TrecQA dataset are shown in Table.~\ref{tab:TrecQA}.
\begin{table}
	\centering
\begin{tabular}{l*{1}{c}}
Method &  Metric=MRR   \\
\hline \\
Re-rank CNN  (\cite{Severyn:2015:}) & 0.8078 \\
Negative Data++  (\cite{HZhang2017}) & 0.8325 \\
D2I & \textbf{0.9061} \\
\end{tabular}
	\caption{Comparative results For TrecQA dataset.}
	\label{tab:TrecQA}
\end{table}

\subsubsection{SICK}
Sentences Involving Compositional Knowledge (SICK) is from Task 1 of the 2014 SemEval competition~(\cite{Marelli}) and consists of 9,927 annotated sentence pairs, with 4,500 for training, 500 as a development set, and 4,927 for testing. Each pair has a relatedness score  [1, 5]
which increases with similarity and an entailment label which takes three values contradict, entail and neutral. The entailment prediction task was a labeling task and we present results for that task only. The comparative results for the SICK dataset are shown in Table.~\ref{tab:SICK}. Note that both the methods DTRNN by \cite{Socher:2011} and DeepPairwise Word by \cite{HeL16} use orders of magnitude more parameters than our proposed model. Also note that since the SICK dataset has both relatedness and entailment signals mixed into the data, \cite{HeL16} propose a KL divergence based cost function. We have maintained the softmax cost function to highlight the generic behavior of the proposed model. Note that with larger amount of data in TwitterPPDB, our model beats the DeepPairwise Word model as shown in Table.~\ref{tab:TPPDBresults}. This points towards over-fitting in case of smaller dataset.
\begin{table}
	\centering
\begin{tabular}{l*{2}{c}}
Method &  & Metric=MSE   \\
\hline \\
DTRNN & \cite{Socher:2011} & 0.3983 \\
DeepPairwiseWord & \cite{HeL16} & \textbf{0.2329} \\
D2I & &  0.3857 \\
\end{tabular}
	\caption{Comparative results For SICK dataset.}
	\label{tab:SICK}
\end{table}






\section{Transfer Learning Experiments}
\subsection{Text to Text Transfer Learning}
We believe that these set of experiments are the first of their kind in this field. 
We start by the simpler experiment, wherein we train a network based on the TwitterPPDB dataset. The hyper-parameter $k=25$ in all these experiments. We choose same network as shown in Fig.~\ref{Fig:CNN}. 
This 5 layer network is trained first with TwitterPPDB data till convergence. We obtain the features from the last dense layer of the trained network, and fine tune it with the TrecQA, WikiQA and the SICK data to generate the results shown in Table.~\ref{tab:TLresults}. Although both the results are lesser than when trained with their own data, the results show that the information transfered through the network is still significant. Also note that for both WikiQA and TrecQA, the transfer learned model still beats the best state-of-the-art models as shown in Table.~\ref{tab:WikiQA} and Table.~\ref{tab:TrecQA}.
\begin{table}
\centering
\begin{tabular}{{c}{c}{c}}
Train Dataset & Fine Tune Dataset & MRR   \\
\hline \\
TwitterPPDB & WikiQA & 0.9758 \\
TwitterPPDB & TrecQA & 0.8589 \\
TwitterPPDB & SICK & 0.7591 \\
\end{tabular}
	\caption{Transfer Learning with TwitterPPDB data and 5 layer CNN network.}
	\label{tab:TLresults}
\end{table}

\subsection{Image to Text Transfer Learning}
We train the same network as shown in Fig.~\ref{Fig:CNN} with MNIST data~(\cite{lecun-mnisthandwrittendigit-2010}) and then fine tune the model with WikiQA, TrecQA and SICK data. We fix the size parameter $k=14$ to match the dimension of MNIST images. This is the first instance of transfer learning between image and text domain. The MRR values for the three datasets are shown in Table.~\ref{tab:CVTLresults}. 

Note that the results are slightly higher than the corresponding values shown in Table.~\ref{tab:TLresults}. The gain in performance may be due to the fact that the base system had better information to train itself by utilizing the MNIST data. The structure present in the MNIST dataset when used for pre-training, renders more generic learning of the weights than achieved by the images generated by the D2I conversion of the Twitter dataset. This leads to better performance for the downstream classification tasks.
These results are extremely interesting, and we propose to continue working on more such experiments in the future.


\begin{table}
\centering
\begin{tabular}{{c}{c}{c}}
Train Dataset & Fine Tune Dataset & MRR   \\
\hline \\
MNIST & WikiQA & 0.9762 \\
MNIST & TrecQA & 0.9064 \\
MNIST & SICK & 0.7600 \\
\end{tabular}
	\caption{Transfer Learning with MNIST data.}
	\label{tab:CVTLresults}
\end{table}

\section{Conclusion and Future Work}
In this paper we have presented a novel yet simple method of incorporating neighborhood information in text and consequently convert it to an image. This representation has some unique properties such as encoding meaningful edge information. This unique representation can now be easily fed to standard networks lifted from the computer vision community. We present comparative results to multiple text classification datasets and show that we are beating most of them. We also present text to text and image to text transfer learning experiments and show that after transforming to the proposed D2I representation, text and vision can be utilized within the ambit of similar models. This is the first attempt at bridging the gap between the separate networks which have been proposed in the vision and text community, and we sincerely hope that this will generate lot of interest within the research community to take this further. The representation is still single channel and hence networks designed for single channel gray scale images can be used for now. Transforming the self-attention images to multiple channels, such that they can exploit more sophisticated models such as Inception-v3~(\cite{Szegedy2016RethinkingTI}) still remains an active research opportunity for the future.


\bibliography{ref}

\end{document}